\documentclass[11pt]{article}
\usepackage{graphicx}
\usepackage{longtable,lscape}
\usepackage{amsthm}
\usepackage{amsfonts}
\usepackage{amsmath}
\usepackage{bbm}
\usepackage{float}
\usepackage{url}
\newtheorem{definition}{Definition}
\newtheorem{theorem}{Theorem}

\newcommand{\captionfonts}{\footnotesize}
\makeatletter  
\long\def\@makecaption#1#2{%
  \vskip\abovecaptionskip
  \sbox\@tempboxa{{\captionfonts #1: #2}}%
  \ifdim \wd\@tempboxa >\hsize
    {\captionfonts #1: #2\par}
  \else
    \hbox to\hsize{\hfil\box\@tempboxa\hfil}%
  \fi
  \vskip\belowcaptionskip}
\makeatother 
\begin{document}
\title{Entanglement Zoo I: Foundational and Structural Aspects}
\author{Diederik Aerts and Sandro Sozzo \vspace{0.5 cm} \\ 
        \normalsize\itshape
        Center Leo Apostel for Interdisciplinary Studies \\
        \normalsize\itshape
        and, Department of Mathematics, Brussels Free University \\ 
        \normalsize\itshape
         Krijgskundestraat 33, 1160 Brussels, Belgium \\
        \normalsize
        E-Mails: \url{diraerts@vub.ac.be,ssozzo@vub.ac.be}
           \\
              }
\date{}
\maketitle
\begin{abstract}
\noindent
We put forward a general classification for a structural description of the entanglement present in compound entities experimentally violating Bell's inequalities, making use of a new entanglement scheme that we developed in \cite{asIQSA2012}. 
Our scheme, although different from the traditional one, is completely compatible with standard quantum theory, and enables quantum modeling in complex Hilbert space for different types of situations. Namely, situations where entangled states and product measurements appear (`customary quantum modeling'), and situations where states and measurements and evolutions between measurements are entangled (`nonlocal box modeling', `nonlocal non-marginal box modeling'). The role played by Tsirelson's bound and marginal distribution law is emphasized. Specific quantum models are worked out in detail in complex Hilbert space within this new entanglement scheme.
\end{abstract}

\section{Introduction\label{intro}}
Entanglement is one of the most intriguing aspects of quantum physics. It is the feature that most neatly marked the departure from ordinary intuition and common sense, on which classical physics rest. The structural and conceptual novelties brought in by quantum entanglement were originally put forward by John Bell in 1964. He proved that, if one introduces `reasonable assumptions for physical theories', one derives an inequality for the expectation values of coincidence measurements performed on compound entities (`Bell's inequality') which does not hold in quantum theory \cite{bell1964}. 
In quantum physics, entanglement is responsible for the violation of this inequality, which entails that quantum particles share statistical correlations that cannot be described in a single classical Kolmogorovian probability framework \cite{af1982,aerts1986,pitowsky1989}. Another 
amazing constatation was that entanglement, together with a number of other quantum features, such as `contextuality', `emergence', `interference' and `superposition', also appears outside the microscopic domain of quantum theory. 
These findings constituted the beginning of a systematic and promising search for quantum structures and the employment of quantum-based models in domains where classical structures prove to be problematical \cite{aertsaerts1995,aertsgabora2005a,aertsgabora2005b,bruzaetal2007,bruzaetal2008,aerts2009,bruzaetal2009,pb2009,k2010,songetal2011,bpft2011,bb2012,busemeyeretal2012}.

As for our own research, many years ago we already identified situations in macroscopic physics which violate Bell's inequalities \cite{aerts1982,aerts1985a,aerts1985b,aerts1991,aertsaertsbroekaertgabora2000}. One of these macroscopic examples, the `connected vessels of water', exhibits even a maximal possible violation of Bell's inequalities, i.e. more than the typical entangled spin example in quantum physics. More recently, we performed a cognitive experiment showing that a specific combination of concepts, {\it The Animal Acts}, violates Bell's inequalities \cite{as2011,ags2012,abgs2012}. These two situations present deep structural and conceptual analogies which we analyze systematically in Ref. \cite{asIQSA2012,asQI2013}. 

In the present paper, we put forward a classification that enables us to represent experimental situations of compound entities which violate Bell's inequalities identifying the quantum-theoretic modeling involved in these violations. We show that a complete quantum-mechanical representation can be worked out, and we prove that quantum entanglement not only appears on the level of the states, but also on the level of the measurements. Indeed, we show that the empirical data we collected on {\it The Animal Acts} (Sec. \ref{technical}), as well as the situation of the `connected vessels of water' \cite{asQI2013}, can be modeled only when both states and measurements are entangled.
The existence of a quantum model for the `connected vessels of water' was not a priori expected and constitutes an original result. Our modeling scheme, although completely compatible with standard quantum theory, is more general than the traditional one, because within this traditional scheme certain ways in which subentities can be part of compound entities have been overlooked. It is when the marginal probability law is violated that this shortcoming of the traditional entanglement scheme comes on the surface, and hence some of the entanglement situations that we consider in the present paper would not be possible to be modeled within the traditional entanglement scheme.

Our classification gives rise to the following different types of situations and entities: {\bf Type 1}: Situations where Bell's inequalities are violated within `Tsirelson's bound' \cite{tsirelson80} and the marginal distribution law holds (`customary quantum modeling'), (Sec. \ref{animalacts}); {\bf Type 2}: Situations where Bell's inequalities are violated within Tsirelson's bound and the marginal distribution law is violated (`nonlocal non-marginal box modeling 1'), (Sec. \ref{animalacts}).
We recall that situations of type 2 seem to be present in `real quantum spin experiments'. A reference to the `experimental anomaly' that, in our opinion, indicates the presence of entangled measurements, occurs already in Alain AspectÕs PhD thesis \cite{aspect1982,ak}.
Our framework accommodates these situations too; {\bf Type 3}: Situations where Bell's inequalities are violated beyond Tsirelson's bound and the marginal distribution law is violated (`nonlocal non-marginal box modeling 2'), (Sec. \ref{animalacts}); {\bf Type 4}: Situations where Bell's inequalities are violated beyond Tsirelson's bound and the marginal distribution law holds (`nonlocal box modeling'), (Sec. \ref{glimmerinkling}).

Additionally to introducing the framework, we analyze in this paper the hypothesis that `satisfying the marginal distribution law' is merely a consequence of extra symmetry being present in situations that contain full-type entanglement, e.g., situations of types 2 and 3. Whenever enough symmetry is present, such that all the entanglement of the situation can be pushed into the state, allowing a model with product measurements, 
and product unitary transformations, the marginal law is satisfied. We give two examples, a cognitive `gedanken experiment' violating Bell's inequalities, which is a `variation adding more symmetry' to an example that was introduced in Ref. \cite{aertsaertsbroekaertgabora2000}, and in this variation the marginal law  is satisfied. We introduce in a similar way extra symmetry in our `vessels of water' example, to come to a variation where the marginal law is satisfied. Both examples are isomorphic and realizations of the so-called `nonlocal box', which is studied as a purely theoretical construct -- no physical realizations were found prior to the ones we present here  -- in the foundations of quantum theory \cite{rp94}.

Let us state clearly, to avoid misunderstandings, that we use the naming `entanglement' referring explicitly to the structure within the theory of quantum physics that a modeling of experimental data take, if (i) these data are represented, following carefully the rules of standard quantum theory, in a complex Hilbert space, and hence states, measurements, and evolutions, are presented respectively by vectors (or density operators), self-adjoint operators, and unitary operators in this Hilbert space; (ii) a situation of coincidence joint measurement on a compound entity is considered, and the subentities are identified following the tensor product rule of `compound entity description in quantum theory' (iii) within this tensor product description of the compound entity entanglement is identified, as `not being product', whether it is for states (non product vectors), measurements (non-product self-adjoint operators), or transformations (non-product unitary transformations). 

\section{Technical aspects of modeling quantum entanglement \label{technical}}
To develop our new scheme for the study of entanglement, we will first introduce some basic notions and results, which we developed in \cite{asIQSA2012} in detail. The more general nature of our scheme, as compared to the standard one, is that we carefully analyse the different ways in which two entities can be subentities of a compound entity. Indeed, entanglement depends crucially on these different possible way of `being a subentity', and this has not been recognised sufficiently in the standard scheme. Let us also remark explicitly that, although our scheme is more general than the standard one, it is completely compatible with standard quantum theory. Hence, the limitations and simplifications as compared to our scheme of the standard one are only linked to an overlooking of the more subtle ways in which subentities can place themselves within a compound entity in situations described by quantum theory.

First we introduce the notions of `product state', `product measurement' and `product dynamical evolution' as we will use it in our new entanglement scheme. For this we consider the general form of an isomorphism $I:{\mathbb C}^4 \rightarrow {\mathbb C}^2 \otimes {\mathbb C}^2$, by linking the elements of an ON basis $\{|x_1\rangle, |x_2\rangle, |x_3\rangle, |x_4\rangle \}$ of ${\mathbb C}^4$ to the elements $\{|c_1\rangle\otimes|d_1\rangle, |c_1\rangle\otimes |d_2\rangle, |c_2\rangle\otimes|d_1\rangle, |c_2\rangle\otimes|d_2\rangle\}$ of the type of ON basis of ${\mathbb C}^2 \otimes {\mathbb C}^2$ where $\{|c_1\rangle, |c_2\rangle\}$ and $\{|d_1\rangle, |d_2\rangle\}$ are ON bases of ${\mathbb C}^2$ each
\begin{equation}
I|x_1\rangle=|c_1\rangle\otimes|d_1\rangle,\ I|x_2\rangle=|c_1\rangle\otimes |d_2\rangle \label{corr1},\ I|x_3\rangle=|c_2\rangle\otimes|d_1\rangle,\ I|x_4\rangle=|c_2\rangle\otimes|d_2\rangle
\end{equation}
\begin{definition}
A state $p$ represented by the unit vector $|p\rangle \in {\mathbb C}^4$ is a `product state', with respect to $I$, if there exists two states $p_a$ and $p_b$, represented by the unit vectors $|p_a\rangle \in {\mathbb C}^2$ and $|p_b\rangle \in {\mathbb C}^2$, respectively, such that $I|p\rangle=|p_a\rangle\otimes|p_b\rangle$. Otherwise, $p$ is an `entangled state' with respect to $I$.
\end{definition}
\begin{definition}
A measurement $e$ represented by a self-adjoint operator ${\cal E}$ in ${\mathbb C}^4$ is a `product measurement', with respect to $I$, if there exists
measurements $e_a$ and $e_b$, represented by the self-adjoint operators  ${\cal E}_a$ and ${\cal E}_b$, respectively, in ${\mathbb C}^2$ such that $I{\cal E}I^{-1}={\cal E}_a \otimes {\cal E}_b$. Otherwise, $e$ is an `entangled measurement' with respect to $I$.
\end{definition}
\begin{definition}
A dynamical evolution $u$ represented by a unitary operator ${\cal U}$ in ${\mathbb C}^4$ is a `product evolution', with respect to $I$, if there exists
dynamical evolutions $u_a$ and $u_b$, represented by the unitary operators operators  ${\cal U}_a$ and ${\cal U}_b$, respectively, in ${\mathbb C}^2$ such that $I{\cal U}I^{-1}={\cal U}_a \otimes {\cal U}_b$. Otherwise, $u$ is an `entangled evolution' with respect to $I$.
\end{definition}
Remark that the notion of product states, measurements and evolutions, are defined with respect to the considered isomorphism between $I:{\mathbb C}^4 \rightarrow {\mathbb C}^2 \otimes {\mathbb C}^2$, which expresses already the new aspect of our entanglement scheme, making entanglement depending on `how sub entities are part of the compound entity'.
The following theorems can then be proved.
\begin{theorem}\label{th1}
The spectral family of a self-adjoint operator ${\cal E}=I^{-1}{\cal E}_a \otimes {\cal E}_bI$ representing a product measurement with respect to $I$, has the form $\{I^{-1}|a_1\rangle \langle a_1|\otimes |b_1\rangle \langle b_1|I$, $I^{-1}|a_1\rangle \langle a_1| \otimes |b_2\rangle \langle b_2|I$, $I^{-1}|a_2\rangle \langle a_2| \otimes |b_1\rangle \langle b_1|I, I^{-1}|a_2\rangle \langle a_2| \otimes |b_2\rangle \langle b_2|I\}$, where $\{|a_1\rangle \langle a_1|, |a_2\rangle \langle a_2|\}$ is a spectral family of ${\cal E}_a$ and $\{|b_1\rangle \langle b_1|, |b_2\rangle \langle b_2|\}$ is a spectral family of ${\cal E}_b$. 
\end{theorem}
Theorem \ref{th1} shows that the spectral family of a product measurement is made up of product orthogonal projection operators.
\begin{theorem} \label{th2}
Let $p$ be a product state represented by the vector $|p\rangle=I^{-1}|p_a\rangle\otimes|p_b\rangle$ with respect to the isomorphism $I$, and $e$ a product measurement represented by the self-adjoint operator ${\cal E}=I{\cal E}_a \otimes {\cal E}_bI^{-1}$ with respect to the same $I$. Let $\{|y_1\rangle, |y_2\rangle, |y_3\rangle, |y_4\rangle\}$ be the ON basis of eigenvectors of ${\cal E}$, and $\{|a_1\rangle, |a_2\rangle\}$ and $\{|b_1\rangle, |b_2\rangle\}$ the ON bases of eigenvectors of ${\cal E}_a$ and ${\cal E}_b$ respectively. Then, we have  $p(A_1)+p(A_2)=p(B_1)+p(B_2)=1$, and  $p(Y_1)=p(A_1)p(B_1)$, $p(Y_2)=p(A_1)p(B_2)$, $p(Y_3)=p(A_2)p(B_1)$ and $p(Y_4)=p(A_2)p(B_2)$, where $\{p(Y_1), p(Y_2),$ $p(Y_3), p(Y_4)\}$ are the probabilities to collapse to states $\{|y_1\rangle, |y_2\rangle, |y_3\rangle, |y_4\rangle\}$, and $\{p(A_1), p(A_2)\}$ and $\{p(B_1), p(B_2)\}$ are the probabilities to collapse to states $\{|a_1\rangle,$ $|a_2\rangle\}$ and $\{|b_1\rangle, |b_2\rangle\}$ respectively.
\end{theorem}
With this theorem we prove that if there exists an isomorphism $I$ between ${\mathbb C}^4$ and ${\mathbb C}^2 \otimes {\mathbb C}^2$ such that state and measurement are both product with respect to this isomorphism, then the probabilities factorize.
A consequence is that in case the probabilities do not factorize the theorem is not satisfied. This means that there does not exist an isomorphism between ${\mathbb C}^4$ and ${\mathbb C}^2 \otimes {\mathbb C}^2$ such that both state and measurement are product with respect to this isomorphism, and there is genuine entanglement. The above theorem however does not yet prove where this entanglement is located, and how it is structured. The next theorems tell us more about this.

We consider now the coincidence measurements $AB$, $AB'$, $A'B$ and $A'B'$ from a typical Bell-type experimental setting. For each measurement we consider the ON bases of its eigenvectors in ${\mathbb C}^4$. For the measurement $AB$ this gives rise to the unit vectors $\{|ab_{11}\rangle, |ab_{12}\rangle, |ab_{21}\rangle, |ab_{22}\rangle\}$, for $AB'$ to the vectors $\{|ab'_{11}\rangle, |ab'_{12}\rangle, |ab'_{21}\rangle, |ab'_{22}\rangle\}$, for $A'B$ to the unit vectors $\{|a'b_{11}\rangle, |a'b_{12}\rangle, |a'b_{21}\rangle,$ $|a'b_{22}\rangle\}$ and for $A'B'$ to the vectors $\{|a'b'_{11}\rangle,|a'b'_{12}\rangle,$ $  |a'b'_{21}\rangle, |a'b'_{22}\rangle\}$. We introduce the dynamical evolutions $u_{AB'AB}$, \ldots, represented by the unitary operators ${\cal U}_{AB'AB}$,\ldots, connecting the different coincidence experiments for any combination of them, i.e. ${\cal U}_{AB'AB}:{\mathbb C}^4 \rightarrow {\mathbb C}^4$, such that  
\begin{equation}
|ab_{11}\rangle \mapsto |ab'_{11}\rangle, |ab_{12}\rangle \mapsto |ab'_{12}\rangle, |ab_{21}\rangle \mapsto |ab'_{21}\rangle, |ab_{22}\rangle \mapsto |ab'_{22}\rangle 
\end{equation}
\begin{theorem}\label{measuremententanglement}
There exists a isomorphism between ${\mathbb C}^4$ and ${\mathbb C}^2 \otimes {\mathbb C}^2$ with respect to which both measurements $AB$ and $AB'$ are product measurements iff there exists an isomorphism between ${\mathbb C}^4$ and ${\mathbb C}^2 \otimes {\mathbb C}^2$ with respect to which the dynamical evolution $u_{AB'AB}$ is a product evolution and one of the measurements is a product measurement. In this case the marginal law is satisfied for the probabilities connected to these measurements, i.e. $p(A_1,B_1)+p(A_1,B_2)=p(A_1,B'_1)+p(A_1,B'_2)$.
\end{theorem}
The above theorem introduces an essential deviation of the customary entanglement scheme, which we had to consider as a consequence of our experimental data on the concept combination {\it The Animal Acts}. Indeed, considering our description of the situation in Section \ref{animalacts}, we have $P(A_1, B_1)+p(A_1,B_2)=0.679\ne0.618=p(A_1,B'_1)+p(A_1,B'_2)$, which shows that the marginal law is not satisfied for our data. Hence, for the our experimental data on {\it The Animal Acts} there does not exist an isomorphism between 
${\mathbb C}^4$ and ${\mathbb C}^2\otimes{\mathbb C}^2$, such that with respect to this isomorphism all measurements that we performed in our experiment can be considered to be product measurements. It right away shows that we will not able to model our data within the customary entanglement scheme.
We could have expected this, since indeed, in this customary scheme all considered measurements are product measurements, and entanglement only appears in the state of the compound entity. We refer to Ref. \cite{asIQSA2012} for proof of Ths. \ref{th1}--\ref{measuremententanglement}.
 
Let us summarise the structural situation. Entanglement is a 
property attributed to states, measurements, or unitary transformations, when looked at the tensor product identification (isomorphism) with the Hilbert space describing the compound entity. The `physics' of the compound entity is expressed in this one Hilbert space describing directly the compound entity, which makes entanglement itself dependent on `the way in which subentities of the compound entity are attempted to be identified'. For one state and one compound measurements, the identification between tensor product and compound entity Hilbert space can always be chosen such that the measurement appears as a product, and all the entanglement is pushed in the state. Theorem \ref{measuremententanglement} 
shows that, whenever the marginal distribution law is violated,
this can no longer be achieved, and entanglement is also present in measurements and the dynamical transformations connecting these measurements. In \cite{asIQSA2012} we show that in case different isomorphisms of identification are considered, an entanglement scheme with again product measurements and product dynamical transformations is possible. But the price to pay is that the entangled state cannot be presented any longer in a unique way within the tensor product space, i.e. a different representation is needed for each coincidence experiment context. All this of course related to the marginal law for the probabilities connected to these different coincidence measurements not being satisfied.
A direct consequence of the above is that, if a set of experimental data violate both Bell's inequalities and the marginal distribution law, it is impossible to work out a quantum-mechanical representation in a fixed Hilbert space ${\mathbb C}^2\otimes{\mathbb C}^2$ which satisfies the data and where only the initial state is entangled while all measurements are products. We will make this more explicit in the next sections. 

\section{Examples of systems entailing entanglement\label{animalacts}}
The first example we shortly present is that of a macroscopic entity violating Bell's inequalities in exactly the same way as a pair of spin-$1/2$ quantum particles in the singlet spin state when faraway spin measurements are performed \cite{aerts1991}. We only sketch this example here to make our zoo collection as complete as possible, and refer to \cite{aerts1991,aertsaertsbroekaertgabora2000} for a detailed presentation.

This mechanical entity simulates the singlet spin state of a pair of spin-$1/2$ quantum particles by means of two point particles $P_1$ and 
$P_2$ initially located in the centers $C_1$ and $C_2$ of two separate unit spheres $B_1$ and $B_2$, respectively. The centers $C_1$ and $C_2$ remain connected by a rigid but extendable rod, which introduces correlations. We denote this state of the overall entity by $p_{s}$. A measurement $A(a)$ is performed on $P_1$ which consists in installing a piece of elastic of 2 units of length between the diametrically opposite points $-a$ and $+a$ of $B_1$. At one point, the elastic breaks somewhere and $P_1$ is drawn toward either $+a$ (outcome $\lambda_{A_{1}}=+1$) or $-a$ (outcome $\lambda_{A_{2}}=-1$). Due to the connection, $P_2$ is drawn toward the opposite side of $B_2$ as compared to $P_1$. Now, an analogous measurement $B(b)$ is performed on $P_2$ which consists in installing a piece of elastic of 2 units of length between the two diametrically opposite points $-b$ and $+b$ of $B_2$. The particle $P_2$ falls onto the elastic following the orthogonal path and sticks there. Next the elastic breaks somewhere and drags $P_2$ toward either $+b$ (outcome $\lambda_{B_{1}}=+1$) or $-b$ (outcome $\lambda_{B_2}=-1$). To calculate the transition probabilities, we assume there is a uniform probability of breaking on the elastics. The single and coincidence probabilities coincide with the standard probabilities for spin-$1/2$ quantum particles in the singlet spin state when spin measurements are performed along directions $a$ and $b$. In particular, the probabilities for the coincidence counts $\lambda_{A_{1}B_{1}}=\lambda_{A_{2}B_{2}}=+1$ and $\lambda_{A_{1}B_{2}}=\lambda_{A_{2}B_{1}}=-1$ of the joint measurement $AB(a,b)$ in the state $p_{s}$ are given by
\begin{eqnarray}
p(p_s,AB(a,b),\lambda_{A_{1}B_{1}})=p(p_s,AB(a,b),\lambda_{A_{2}B_{2}})={1 \over 2} \sin^{2}{\gamma \over 2} \\
p(p_s,AB(a,b),\lambda_{A_{1}B_{2}})=p(p_s,AB(a,b),\lambda_{A_{2}B_{1}})={1 \over 2} \cos^{2}{\gamma \over 2}
\end{eqnarray}
respectively, where $\gamma$ is the angle between $a$ and $b$, in exact accordance with the quantum-mechanical predictions. Furthermore, this model leads to the same violation of Bell's inequalities as standard quantum theory. Hence, the `connected spheres model' is structurally isomorphic to a standard quantum entity. This means that it can be represented in the Hilbert space ${\mathbb C}^{2}\otimes{\mathbb C}^{2}$ in such a way that its initial state is the singlet spin, i.e. a maximally entangled state, and the measurements are products. Furthermore, the marginal distribution law holds and Bell's inequalities are violated within the Tsirelson's bound $2\sqrt{2}$, hence the connected spheres model is an example of a `customary identified standard quantum modeling' in our theoretical framework.

The presence of entanglement in concept combination has recently also been identified in a cognitive test \cite{as2011,ags2012,abgs2012} and subsequently improved by elaborating a quantum Hilbert space modeling of it \cite{asIQSA2012,asQI2013}. We analyze it in the light of our 
our new entanglement scheme exposed in Sec. \ref{technical}.
For a detailed description of the conceptual entity, and the measurements considered, we refer to \cite{asQI2013}, Sec. 2.1,
or \cite{asIQSA2012}.
We consider the typical Bell inequality situation of four coincidence measurements $AB$, $AB'$, $A'B$ and $A'B'$, performed on the sentence {\it The Animal Acts} as a conceptual combination of the concepts {\it Animal} and {\it Acts}. Measurements consists of asking participants in the experiment to answer the question whether a given exemplar `is a good example' of the considered concept or conceptual combination.

We had 81 subjects participating in our experiment. If we denote by $p(A_1,B_1)$, $p(A_1,B_2)$, $p(A_2,B_1)$, $p(A_2,B_2)$ the probabilities for the different Bell-type situation choices, we find
$p(A_1,B_1)=0.049$, $p(A_1,B_2)=0.630$, $p(A_2,B_1)=0.259$, $p(A_2,B_2)=0.062$, $p(A_1,B'_1)=0.593$, $p(A_1,B'_2)=0.025$, $p(A_2,B'_1)=0.296$, $p(A_2,B'_2)=0.086$, $p(A'_1,B_1)=0.778$, $p(A'_1,B_2)=0.086$, $p(A'_2,B_1)=0.086$, $p(A'_2,B_2)=0.049$, $p(A'_1,B'_1)=0.148$, $p(A'_1,B'_2)=0.086$, $p(A'_2,B'_1)=0.099$, $p(A'_2,B'_2)=0.667$, and the expectation values are $E(A,B)=p(A_1,B_1)-p(A_1,B_2)-p(A_2,B_1)+p(A_2,B_2)=-0.7778$, $E(A,B')=p(A_1,B'_1)-p(A_1,B'_2)-p(A_2,B'_1)+p(A_2,B'_2)=0.3580$, $E(A',B)=p(A'_1,B_1)-p(A'_1,B_2)-p(A'_2,B_1)+p(A'_2,B_2)=0.6543$,  $E(A',B')=p(A'_1,B'_1)-p(A'_1,B'_2)-p(A'_2,B'_1)+p(A'_2,B'_2)=0.6296$.
Inserting them into the Clauser-Horne-Shimony-Holt (CHSH) version of Bell's inequality \cite{chsh69} 
\begin{equation} \label{chsh}
-2 \le E(A',B')+E(A',B)+E(A,B')-E(A,B) \le 2.
\end{equation}
we find
$E(A',B')+E(A',B)+E(A,B')-E(A,B)=2.4197$. This violation proves  the presence of entanglement in the conceptual situation considered.

The probabilities corresponding to the coincidence measurements cannot be factorized, which means that a result stronger than the one in Th. \ref{th2} holds. For example, for the measurement $AB$, there do not exist real numbers $a_1, a_2, b_1, b_2\in [0,1]$, $a_1+a_2=1$, $b_1+b_2=1$, such that $a_1b_1=0.05$, $a_2b_1=0.63$, $a_1b_2=0.26$ and $a_2b_2=0.06$. Indeed, supposing that such numbers do exist, from $a_2b_1=0.63$ follows that $(1-a_1)b_1=0.63$, and hence $a_1b_1=1-0.63=0.37$. This is in contradiction with $a_1b_1=0.05$. It is also easy to verify that that marginal law is not satisfied, for example $p(A'_1,B_1)+p(A'_1,B_2)=0.864\not=0.234=p(A'_1,B'_1)+p(A'_1,B'_2)$.
Following theorem \ref{measuremententanglement} a quantum representation where only the state is entangled, while all measurements are products, does not exist. But a representation which entails entangled measurements can be elaborated \cite{asIQSA2012,asQI2013}. 
In the quantum modeling we worked out, the state of {\it The Animal Acts} is represented by a non-maximally entangled state, while all coincidence measurements are entangled. Since the violation of the CHSH inequality we found satisfies Tsirelson's bound, this quantum modeling for the concept combination {\it The Animal Acts} is an example of a `nonlocal non-marginal box modeling 1'.

Next we consider the `vessels of water' example \cite{aerts1982,aerts1985a,aerts1985b}. Two vessels $V_A$ and $V_B$ are interconnected by a tube $T$, vessels and tube containing 20 liters of transparent water. The measurements $A$ and $B$ consist in siphons $S_A$ and $S_B$ pouring out water from vessels $V_A$ and $V_B$, respectively, and collecting the water in reference vessels $R_A$ and $R_B$, where the volume of collected water is measured. If more than 10 liters are collected for $A$ or $B$, we put $\lambda_{A_{1}}=+1$ or $\lambda_{B_{1}}=+1$, respectively, and if fewer than 10 liters are collected for $A$ or $B$, we put $\lambda_{A_{2}}=-1$ or $\lambda_{B_{2}}=-1$, respectively. Measurements $A'$ and $B'$ consist in taking a small spoonful of water out of the left vessel and the right vessel, respectively, and verifying whether the water is transparent. We have $\lambda_{A'_{1}}=+1$ or $\lambda_{A'_{2}}=-1$, depending on whether the water in the left vessel turns out to be transparent or not, and 
$\lambda_{B'_{1}}=+1$ or $\lambda_{B'_{2}}=-1$, depending on whether the water in the right vessel turns out to be transparent or not. We put $\lambda_{A_{1}B_{1}}=\lambda_{A_{2}B_{2}}=+1$ if $\lambda_{A_{1}}=+1$ and $\lambda_{B_{1}}=+1$ or $\lambda_{A_{2}}=-1$ and $\lambda_{B_{2}}=-1$, and $\lambda_{A_{1}B_{2}}=\lambda_{A_{2}B_{1}}=-1$ if $\lambda_{A_{1}}=+1$ and $\lambda_{B_{2}}=-1$ or $\lambda_{A_{2}}=-1$ and $\lambda_{B_{1}}=+1$, if the coincidence measurement $AB$ is performed. We proceed analogously for the outcomes of the measurements $AB'$, $A'B$ and $A'B'$. We can then define the expectation values $E(A,B)$, $E(A,B')$, $E(A',B)$ and $E(A',B')$ associated with these coincidence measurements. Since each vessel contains 10 liters of transparent water, we find that $E(A,B)=-1$, $E(A',B)=+1$, $E(A,B')=+1$ and $E(A',B')=+1$, which gives
$E(A',B')+E(A',B)+E(A,B')-E(A,B)=+4$.
This is the maximal violation of the CHSH inequality and it obviously exceeds Tsirelson's bound. We further have $0.5=p(\lambda_{A_{1}B_{1}})+p(\lambda_{A_{1}B_{2}}) \ne p(\lambda_{A_{1}B'_{1}})+p(\lambda_{A_{1}B'_{2}})=1$,
which shows that the marginal distribution law  is 
violated. In \cite{asQI2013} we constructed a quantum model in complex Hilbert space for the vessels of water situation, where the state $p$ with transparent water and the state $q$ with non-transparent water are entangled, and the measurement $AB$, since it has product states in its spectral decomposition, is a product measurement (Th. \ref{th1}). Compatible with theorem \ref{measuremententanglement} we can see that $AB'$, $A'B$ and $A'B'$ are entangled measurements. Summarizing, we can say that the `vessels of water' situation is an example of a `nonlocal non-marginal box modeling 2'.

\section{Nonlocal boxes\label{glimmerinkling}}
We conclude this paper by giving two examples, the one physical and the other cognitive, which maximally violate Bell's inequalities, i.e. with value 4, but satisfy the marginal distribution law. These examples are also inspired by the macroscopic non-local box example worked out already in 2005 by Sven Aerts, using a breakable elastic and well defined experiments on this elastic \cite{aerts2005}. In physics, a system that behaves in this way is called a `nonlocal box' \cite{rp94}.

For the first example, we again consider the vessels of water and two measurements for each side $A$ and $B$.
The first consists in using the siphon and checking the water. If there are more than 10 liters and the water is transparent ($\lambda_{A_{1}B_{1}}$) or if there are fewer than 10 liters and the water is not transparent ($\lambda_{A_{2}B_{2}}$), the outcome of the first measurement is $+1$. In case there are fewer than 10 liters and the water is transparent $\lambda_{A_{2}B_{1}}$, or if there are more than 10 liters and the water is not transparent $\lambda_{A_{1}B_{2}}$, the outcome is $-1$. The second measurement consists in taking out some water with a little spoon to see if it is transparent or not; if it is transparent, the outcome is $\lambda_{A_{1}B'_{1}}=\lambda_{A_{2}B'_{2}}=+1$, and if it is not transparent, the outcome is $\lambda_{A_{2}B'_{1}}=\lambda_{A_{1}B'_{2}}=-1$.
The water is prepared in a mixed state $m$ of the states $p$ (transparent water) and $q$ (not transparent water) with equal weights. Thus, $m$ is represented by the density operator $\rho=0.5|p\rangle\langle p|+0.5|q\rangle\langle q|$, where $|p\rangle=|0,\sqrt{0.5}e^{i\alpha},0.5e^{i\beta},0 \rangle$ and $|q\rangle=|0,\sqrt{0.5}e^{i\alpha},-0.5e^{i\beta},0 \rangle$ \cite{asQI2013}. 

The coincidence measurement $AB$ is represented by the ON set
$|r_{A_{1}B_{1}}\rangle=|1, 0, 0, 0\rangle$, 
$|r_{A_{1}B_{2}}\rangle=|0, 1, 0, 0\rangle$,
$|r_{A_{2}B_{1}}\rangle=|0, 0, 1, 0\rangle$,
$|r_{A_{2}B_{2}}\rangle=|0, 0, 0, 1\rangle$,
which gives rise to a self-adjoint operator
\small
\begin{equation}
{\mathcal E}_{AB}=\left( \begin{array}{cccc}
\lambda_{A_{1}B_{1}} & 0 & 0 & 0 \\
0 & \lambda_{A_{1}B_{2}} & 0 & 0 \\
0 & 0 & \lambda_{A_{2}B_{1}} & 0 \\
0 & 0 & 0 & \lambda_{A_{2}B_{2}}
\end{array} \right)
\end{equation}
\normalsize
Applying L\"{u}ders' rule, we calculate the density operator representing the state after $AB$. This gives
\begin{equation}
\rho_{AB}=\sum_{i,j=1}^{2} |r_{A_{i}B_{j}}\rangle\langle r_{A_{i}B_{j}}|\rho|r_{A_{i}B_{j}}\rangle\langle r_{A_{i}B_{j}}|=\rho
\end{equation}
as one can easily verify. This means that the nonselective measurement $AB$ leaves the state $m$ unchanged or, equivalently, the marginal distribution law holds.

Measurement $AB'$ is represented by the ON set
$|r_{A_{1}B'_{1}}\rangle=|0, \sqrt{0.5}e^{i\alpha}, \sqrt{0.5}e^{i\beta}, 0\rangle$,
$|r_{A_{1}B'_{2}}\rangle= | 1, 0, 0, 0\rangle$,
$|r_{A_{2}B'_{1}}\rangle= |0, 0, 0, 1\rangle$,
$|r_{A_{2}B'_{2}}\rangle= |0, \sqrt{0.5}e^{i\alpha}, -\sqrt{0.5}e^{i\beta}, 0\rangle$,  
which gives rise to a self-adjoint operator
\small
\begin{equation}
{\mathcal E}_{AB'}=\left( \begin{array}{cccc}
\lambda_{A_{1}B'_{2}} & 0 & 0 & 0 \\
0 & 0.5(\lambda_{A_{1}B'_{1}}+\lambda_{A_{2}B'_{2}}) & 0.5e^{i(\alpha-\beta)}(\lambda_{A_{1}B'_{1}}-\lambda_{A_{2}B'_{2}}) & 0 \\
0 & 0.5e^{-i(\alpha-\beta)}(\lambda_{A_{1}B'_{1}}-\lambda_{A_{2}B'_{2}}) & 0.5(\lambda_{A_{1}B'_{1}}+\lambda_{A_{2}B'_{2}}) & 0 \\
0 & 0 & 0 & \lambda_{A_{2}B'_{1}}
\end{array} \right)
\end{equation}
\normalsize
Applying L\"{u}ders' rule, we calculate the density operator representing the state oafter $AB'$, which gives
\begin{equation}
\rho_{AB'}=\sum_{i,j=1}^{2} |r_{A_{i}B'_{j}}\rangle\langle r_{A_{i}B'_{j}}|\rho|r_{A_{i}B'_{j}}\rangle\langle r_{A_{i}B'_{j}}|=\rho
\end{equation}
Also in this case, the nonselective measurement $AB'$ leaves the state $m$ unchanged. The measurements $A'B$ and $A'B'$ are analogous to $AB'$, hence
 the marginal distribution law is always satisfied. 

We now calculate the expectation values corresponding to the four measurements above in the mixed state $m$ and insert them into the CHSH inequality. This gives
\small
\begin{equation}
{\mathcal E}_{AB}=
\left( \begin{array}{cccc}
1 & 0 & 0 & 0 \\
0 & -1 & 0 & 0 \\
0 & 0 & -1 & 0 \\
0 & 0 & 0 & 1
\end{array} \right)
\quad
{\mathcal E}_{AB'}={\mathcal E}_{A'B}={\mathcal E}_{A'B'}=
\left( \begin{array}{cccc}
-1 & 0 & 0 & 0 \\
0 & 1 & 0 & 0 \\
0 & 0 & 1 & 0 \\
0 & 0 & 0 & -1
\end{array} \right)
\end{equation}
\begin{equation}
B={\mathcal E}_{A'B'}+{\mathcal E}_{A'B}+{\mathcal E}_{AB'}-{\mathcal E}_{AB}=
\left( \begin{array}{cccc}
-4 & 0 & 0 & 0 \\
0 & 4 & 0 & 0 \\
0 & 0 & 4 & 0 \\
0 & 0 & 0 & -4
\end{array} \right)
\end{equation}
\normalsize
Hence, the CHSH inequality
$tr \rho B = 4$,
which shows that Bell inequalities are maximally violated in the mixed state $m$. This construction of a Hilbert space modeling for the `connected vessels of water' is new and was not expected when the original example was conceived.

Next we look at the cognitive example. We consider the concept {\it Cat} and two concrete exemplars of it, called {\it Glimmer} and {\it Inkling}, the names of two brother cats that lived in our research center \cite{aertsaertsbroekaertgabora2000}. The concept {\it Cat} is abstractly described by the state $p$. The experiments we consider are realizing physical contexts that influence the collapse of the concept {\it Cat} to one of its exemplars, or states, {\it Glimmer} or {\it Inkling}, inside the mind of a person being confronted with the physical contexts. It is a `gedanken experiment', in the sense that we put forward plausible outcomes for it, taking into account the nature of the physical contexts, and Liane, the owner of both cats, playing the role of the person. 
We also suppose that Liane sometimes puts a collar with a little bell around the necks of both cats, the probability of this happening being equal to $1/2$. We also suppose that if she does, she always puts them around the necks of both cats. 

The measurement $A$ consists in `{\it Glimmer} appearing in front of Liane as a physical context'. We consider outcome $\lambda_{A_{1}}$ to occur if Liane thinks of {\it Glimmer} and there is a bell, or if she thinks of {\it Inkling} and there is no bell, while outcome $\lambda_{A_{2}}$ occurs if Liane thinks of {\it Inkling} and there is a bell, or if she thinks of {\it Glimmer} and there is no bell. The measurement $B$ consists in `{\it Inkling} appearing in front of Liane as a physical context'. We consider outcome $\lambda_{B_{1}}$ to occur if Liane thinks of {\it Inkling} and there is a bell, or if she thinks of {\it Glimmer} and there is no bell, while outcome $\lambda_{B_{2}}$ occurs if Liane thinks of {\it Glimmer} and there is a bell, or if she thinks of {\it Inkling} and there is no bell. 
Experiment $A'$ consists in `{\it Inkling} appearing in front of Liane as a physical context', and outcome $\lambda_{A'_{1}}$ occurs if Inkling wears a bell, and outcome $\lambda_{A'_{2}}$, if Inkling does not. Experiment $B'$ consists in `{\it Glimmer} appearing in front of Liane as a physical context', and outcome $\lambda_{B'_{1}}$ occurs if Glimmer wears a bell, outcome $\lambda_{B'_{1}}$, if Glimmer does not.

The measurement $AB$ consists in both cats showing up as physical context. Because of the symmetry of the situation, it is plausible to suppose probability $1/2$ that Liane thinks of {\it Glimmer} and probability $1/2$ that she thinks of {\it Inkling}, however, they are mutually exclusive. Also, since both cats either wear bells or do not wear bells, $AB$ produces strict anti-correlation, probability $1/2$ for outcome $\lambda_{A_{1}B_{2}}$ and probability $1/2$ for outcome $\lambda_{A_{2}B_{1}}$. Hence $p(\lambda_{A_{1}B_{2}})=p(\lambda_{A_{2}B_{1}})=1/2$ and $p(\lambda_{A_{1}B_{1}})=p(\lambda_{A_{2}B_{2}})=0$, which gives $E(A, B)=-1$.
The measurement $AB'$ consists in {\it Glimmer} showing up as a physical context. This gives rise to a perfect correlation, outcome $\lambda_{A_{1}B'_{1}}$ or outcome $\lambda_{A_{2}B'_{2}}$, depending on whether {\it Glimmer} wears a bell or not, hence both with probability $1/2$. As a consequence, we have $p(\lambda_{A_{1}B'_{1}})=p(\lambda_{A_{2}B'_{2}})=1/2$ and $p(\lambda_{A_{1}B'_{2}})=p(\lambda_{A_{2}B'_{1}})=0$, and $E(A, B')=+1$. The measurement $A'B$ consists in {\it Inkling} showing up as a physical context, again giving rise to a perfect correlation, outcome $\lambda_{A'_{1}B_{1}}$ or outcome $\lambda_{A'_{2}B_{2}}$, depending on whether {\it Inkling} wears a bell or not, hence both with probability $1/2$. This gives $p(\lambda_{A'_{1}B_{1}})=p(\lambda_{A'_{2}B_{2}})=1/2$ and $p(\lambda_{A'_{1}B_{2}})=p(\lambda_{A'_{2}B_{1}})=0$ and $E(A',B)=+1$. The measurement $A'B'$ consists in both cats showing up as physical 
context, giving rise to a perfect correlation, outcome $\lambda_{A'_{1}B'_{1}}$ or outcome $\lambda_{A'_{2}B'_{2}}$, depending on whether both wear bells or not, hence both with probability $1/2$. This gives $p(\lambda_{A'_{1}B'_{1}})=p(\lambda_{A'_{2}B'_{2}})=1/2$ and $p(\lambda_{A'_{1}B'_{2}})=p(\lambda_{A'_{2}B'_{1}})=0$ and $E(A',B')=+1$.

We find $E(A',B')+E(A',B)+E(A,B')-E(A,B)=4$ in the CHSH inequalitiy. The marginal distribution law is satisfied here, because, e.g., $p(\lambda_{A_{1}B_{1}})+p(\lambda_{A_{1}B_{2}})=p(\lambda_{A_{1}B'_{1}})+p(\lambda_{A_{1}B'_{2}})=1/2$. It is easy to check that the marginal distribution law globally holds in this case. 

The two examples above are structurally isomorphic, i.e. one can provide the same quantum Hilbert space model for both of them. Moreover, they are realizations of what quantum foundations physicists call a `nonlocal box', that is, systems obeying the marginal distribution law but violating Bell's inequalities maximally \cite{rp94}. Following our classification, we call this modeling a `nonlocal box modeling'. The above examples show that it is possible to realise nonlocal boxes in nature and elaborate a Hilbert space modeling for them, contrary to what is usually believed in quantum foundation circles.

\end{document}